\documentclass[letterpaper, 10 pt, conference]{ieeeconf}  

\IEEEoverridecommandlockouts                              

\usepackage{graphicx}
\title{\LARGE \bf Underactuated Biomimetic Autonomous Underwater Vehicle for Ecosystem Monitoring}

\author{Kaustubh Singh \qquad Shivam Kumar \qquad Shreyas Kannan \qquad Mangesh Singh\\ Shashikant Pawar \qquad Sandeep Manjanna
\thanks{All the authors are at Plaksha University, Mohali, India.
        {\tt\small msandeep.sjce@gmail.com}}%
}

\begin{document}

\maketitle
\thispagestyle{empty}
\pagestyle{empty}

\begin{abstract}

In this paper we present an underactuated biomimetic underwater robot that is suitable for ecosystem monitoring in both marine and freshwater environments. We present an updated mechanical design for a fish-like robot and propose minimal actuation behaviors learned using reinforcement learning techniques. We present our preliminary mechanical design of the tail oscillation mechanism and illustrate the swimming behaviors on FishGym simulator, where the reinforcement learning techniques will be tested on.

\end{abstract}

\section{INTRODUCTION}

Recent years have seen growing interest in underwater exploration for ecosystem monitoring, marine education, navigation and rescue. Bio-inspired soft robots, particularly fish-like ones, are well suited for observing marine ecosystems that are fragile and undisturbed. Consequently, there's a concerted effort to explore and develop new propulsion mechanisms. One particularly promising direction of research is drawing inspiration from fish swimming, as fish outperform traditional propellers in both energy efficiency and maneuverability~\cite{govardhan2011fluid}. Their superior performance owes not only to their low drag producing streamlined shape but also, crucially, to the unique motion that enables thrust generation with minimal energy dissipation. Biomimicry-based robotic mobility represents a subfield within bio-inspired design, which seeks to apply natural principles to the engineering of real-world systems. Given the increasing interest in this domain, this study aims to assess the performance of a biomimicry-based propulsion system for underwater robots for efficient swimming and thrust generation. The system is not only designed to emulate the physical attributes of the biological counterpart but also to mimic the maneuverability and control strategies, thus striving to achieve the most authentic portrayal of ‘fish-like swimming’.

However, training these robots to mimic or surpass biological counterparts faces challenges due to limited datasets and high experimentation costs. Simulation has emerged as a vital tool for learning robotic navigation. We propose a novel control mechanism using a single motor for overall flapping motion, enhancing efficiency. It aims to test the model’s performance in controlling pose, following paths, and navigating efficiently underwater, demonstrating the feasibility of the approach for various tasks. A substantial body of research in autonomous underwater vehicles (AUVs), simulation software, and under-actuated AUVs serves as an inspiration for this work. The FishGym simulator, proposed by Liu et al.~\cite{liu2022fishgym} enables the training of reinforcement learning algorithms on a fish-like body with realistic fluid physics modeling. Some recent work on using reinforcement learning techniques for marine vehicle navigation address the challenge of navigating marine environments with under-actuated vehicles~\cite{qu2023deep, chen2023deep}. Unlike them, we  focus on biomimetic fish-like robot with activating only one of the spinal joints for control.

\section{APPROACH}

\subsection{Mechanical Design}

In the past, researchers have been employing off-the-shelf actuators such as motors, pumps, and relatively newer intelligent materials to construct various biomimetic robots. Taking inspiration from Berg et al.~\cite{van2022openfish}, our design incorporates a single-motor cable-driven oscillating system, paired with a passive compliant tail segment, however, it has a different mechanism employed for motion transmission and alternate pulling and stretching of the two cables used in achieving the oscillatory motion mimicking the carangiform swimming. Carangiform mode of swimming is essentially based on the lift force generated by one third of the body, including the airfoil like cross-section of the fish tail, as it moves laterally through the water~\cite{duraisamy2019design}.

\begin{figure}[h]
    \centering
    \includegraphics[width=0.9\columnwidth]{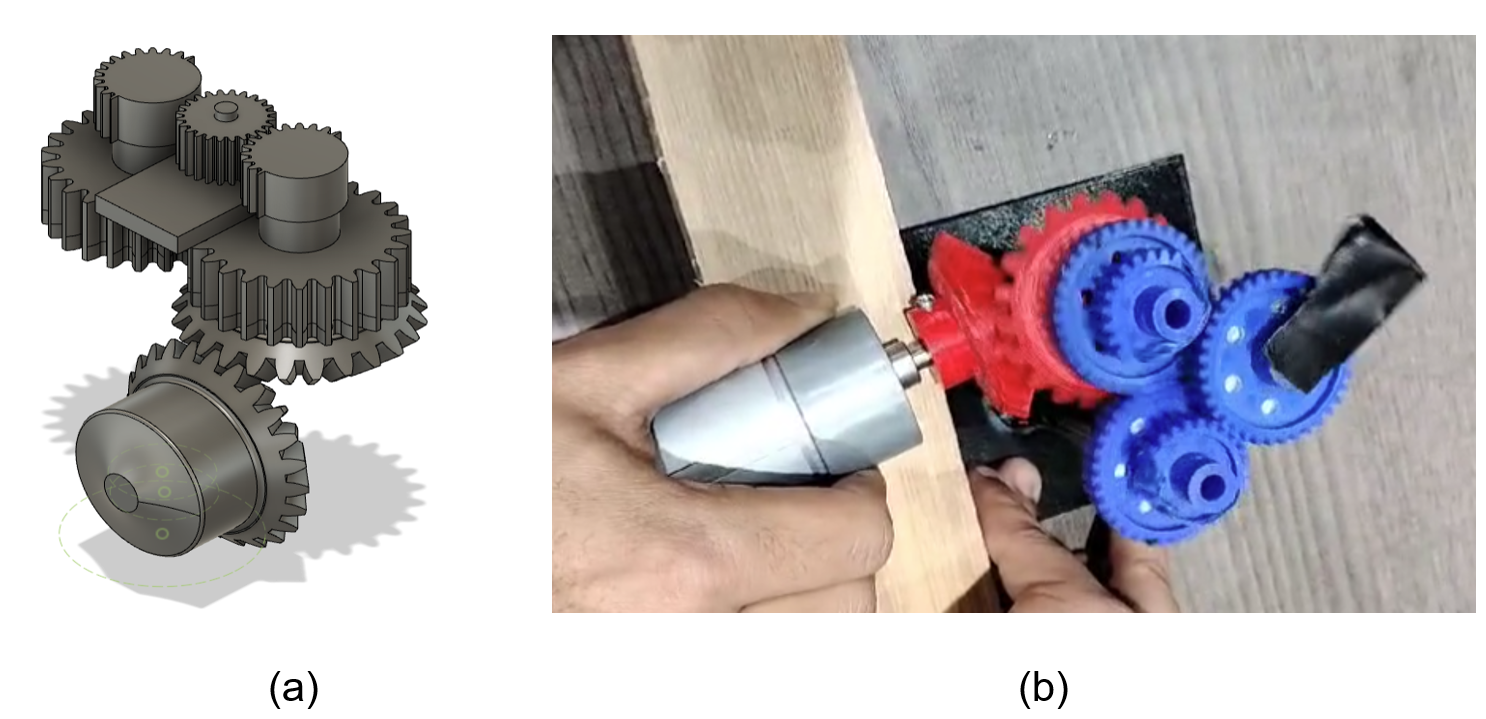}
    \caption{(a) Design of the tail oscillation mechanism and its initial prototype is presented in (b)}\vspace{-2.5pt}
    \label{fig:mech_design}
\end{figure}

A pair of bevel gears is used to transmit the motion at $90^0$, eliminating the need of using the gearbox. As can be seen in Fig.~\ref{fig:mech_design}(a), one of the spur gears is driven by the bevel gear shaft. The oscillatory motion is achieved by an idle gear driven in clockwise and anticlockwise direction alternately by the smaller spur gears with partially shaved teeth. The DC motor's continuous rotation pulls the cables attached to each side of the active tail segment, resulting in a smoother sinusoidal waveform with enhanced speed and efficiency. The active part of the tail is made up of four rigid discs connected by three compliant joints. The stiffness of the passive tail segment is chosen such that an S-shaped tail is realized.

\subsection{Learning to Navigate}

Current underwater robotic systems often lack the agility and efficiency exhibited by biological counterparts. This project
addresses this gap by developing a simulation model that harnesses voltage and current inputs to precisely control tail motion and direction changes in a robotic fish. By emulating the nuanced movements of real fish, the simulated robot will be capable of navigating through complex underwater environments autonomously. Our system is not only designed to emulate the physical attributes of the biological fish but also mimic the maneuverability and control strategies, thus striving to achieve the most authentic portrayal of ‘fish-like swimming’. We plan to achieve this by reducing the control parameters and applying reinforcement learning techniques to learn the controls.

The literature suggests multiple ways of dealing with expansive action spaces. This aspect has been explored in video games~\cite{kanervisto2020action} and they conclude with the options of reducing the number of actions (RA), converting multidiscrete actions to discrete actions (CMD), or discretizing continuous action spaces (DC). Although removing actions can lower performance in some cases, but it can facilitate learnability for the agent in certain game environments. Discretizing continuous actions improves performance in most cases. We have currently reduced the actions from 4 joints being actuated to a single joint control. Reducing actions is of importance in this case as it emulates the physical prototype presented by Berg et. al.~\cite{van2022openfish}.

In FishGym~\cite{liu2022fishgym}, there is a myriad of fish skeletal structures to test out the performance of the simulations. These can be broken down into three main categories - Koi fish, Flat fish, and Eel fish. Moreover, there are clearer and deeper design choices with Koi fish. For example, the setup provides for Koi fish skeletal structure with no active fins ($7$ joints with $4$ active joints), or all possible active fins ($28$ joints with $10$ active joints). For the sake of simplicity, we chose to work with inactive fins for Koi fish. This structure has a total of $7$ joints, namely -$head$, $spine$, $spine01$, $spine02$, $spine03$, $tail\_start$, and $tail\_end$. Out of these $7$ joints, the $4$ spine joints are actuated with force in FishGym. We have made a replica of the design with $3$ of the spine joints as passive joints, only actuating the far end joint of the spine. From the mechanical point of view, this is favorable, as this emulates the movement of the far end of the metal/plastic strip under any pulling motion.

\section{EXPERIMENTS}

We plan to validate the mechanical design through experiments in a controlled environment to measure the thrust generated by the undulatory motion of the tail in water at different frequencies. The driving mechanism will be mounted above the water surface and the tail immersed in water will be connected to the driving mechanism by a hollow shaft.

\begin{figure}[t]
    \centering
    \includegraphics[width=0.6\columnwidth]{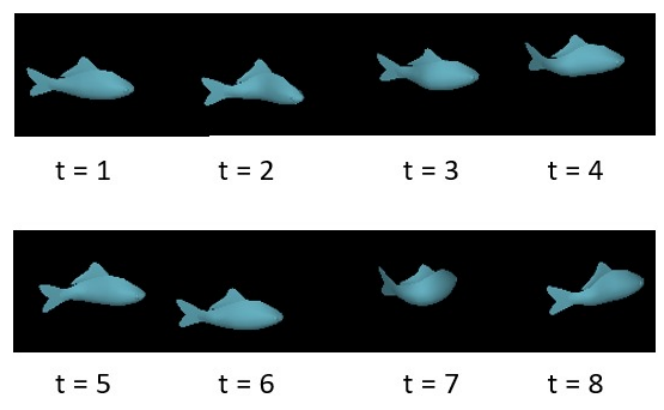}
    \caption{FishGym simulation illustrating various poses of the fish.}\vspace{-5.5pt}
    \label{fig:fishgym}
\end{figure}

The validation for the navigation system will have three evaluation metrics to assess the performance of the reinforcement learning algorithms in the FishGym environment -
\begin{itemize}
    \item \emph{Cruising Efficiency}: This metric assesses how well the robot can swiftly and efficiently reach a specified location. The fish agent will smoothly try to reach the destination, optimizing each movement for maximum efficiency.
    \item \emph{Path Precision}: Path following is a primitive task for robot control. Here, we compare the similarity and difference of paths followed by the agent and FishGym’s intended input path. This metric measures the robot’s ability to follow a given arbitrary path as closely and efficiently as possible.
    \item \emph{Pose Control}: The pose control metric focuses on how well the robot can move and control its body. This includes taking turns, rotating, and adjusting orientation of the agent fish.
\end{itemize}

\section{CONCLUSION}
We presented an underactuated biomimetic robot design and proposed a machine learning based control mechanism to efficiently navigate in marine environments for monitoring natural ecosystems. We propose to reduce the control points on the spine joints of the fish robot from four to just one. Thus making it easier for a learning algorithm to produce good behaviors.
 
\bibliographystyle{IEEEtran}
\bibliography{ref.bib}

\addtolength{\textheight}{-12cm}   




\end{document}